\documentclass{article}

\usepackage[preprint,nonatbib]{neurips_2023}

\usepackage[utf8]{inputenc} 
\usepackage[T1]{fontenc}    
\usepackage{hyperref}       
\usepackage{url}            
\usepackage{booktabs}       
\usepackage{amsfonts}       
\usepackage{nicefrac}       
\usepackage{microtype}      
\usepackage{xcolor}         
\usepackage{subcaption}     
\usepackage[pdftex]{graphicx}
\usepackage[numbers]{natbib}

\title{ArchiGuesser - AI Art Architecture Educational Game}

\author{%
  Joern Ploennigs, Markus Berger, and Eva Carnein \\
  AI for Sustainable Construction\\
  University of Rostock\\
  Rostock, 18059, Germany \\
  \texttt{\{Joern.Ploennigs; Markus.Berger; Eva.Carnein\}@uni-rostock.de} \\
}

\begin{document}

\maketitle

\begin{abstract}
The use of generative AI in education is a controversial topic. Current technology offers the potential to create educational content from text, speech, to images based on simple input prompts. This can enhance productivity by summarizing knowledge and improving communication, quickly adjusting to different types of learners.
Moreover, generative AI holds the promise of making the learning itself more fun, by responding to user inputs and dynamically generating high-quality creative material.

In this paper we present the multisensory educational game \textit{ArchiGuesser} that combines various AI technologies from large language models, image generation, to computer vision to serve a single purpose: Teaching students in a playful way the diversity of our architectural history and how generative AI works.
\end{abstract}

\section{Introduction}
The new challenges of generative AI in education are widely discussed from pros and cons \cite{baidoo2023education,qadir2023engineering,rudolph2023chatgpt} to regulatory implications \cite{hacker2023regulating,sullivan2023chatgpt}. The discussion is dominated by the potential threats of student misuse. 
Practical examples that actually demonstrate how generative AI can be used in education are rare in the discussion with a few examples from teaching code with Chat GPT \cite{surameery2023use} to engaging students' interest in art in STEAM classes with image-generation \cite{lee2023prompt}.
In consequence, many practitioners are already ahead of education in adopting generative AI. A recent study analysing 85 millions of Midjourney prompts has shown that more than 6\,\% are related to architecture \cite{ploennigs2023ai}. The analysis points out that the platform actually has good understanding of architectural styles and people query and combine them across periods and geographic locations to create new designs.

Within this paper we show how we can utilize that in education. We present a multisensory educational game that is combining multiple generative AI technologies to engage students in learning about architectural styles. With the game we particularly put focus on the diversity of architectural history across the globe.
The game principle is simple: Display an image or read a poem for a randomly selected architectural style and encourage the user to make educated guesses about the specific architectural style presented, its geographical origin, and the time period it was created in. The advanced game mode incorporates images and text inspired by real landmarks known for their striking representation of certain architectural styles, adding an extra layer of complexity in identify both the original and newly fused styles, along with their respective locations and time periods. This challenge enhances the gameplay and breaks down traditional cultural boundaries in architectural history. 

\begin{figure}
\subcaptionbox{Gameplay \label{fig:gameplay}}{\includegraphics[width=0.29\textwidth]{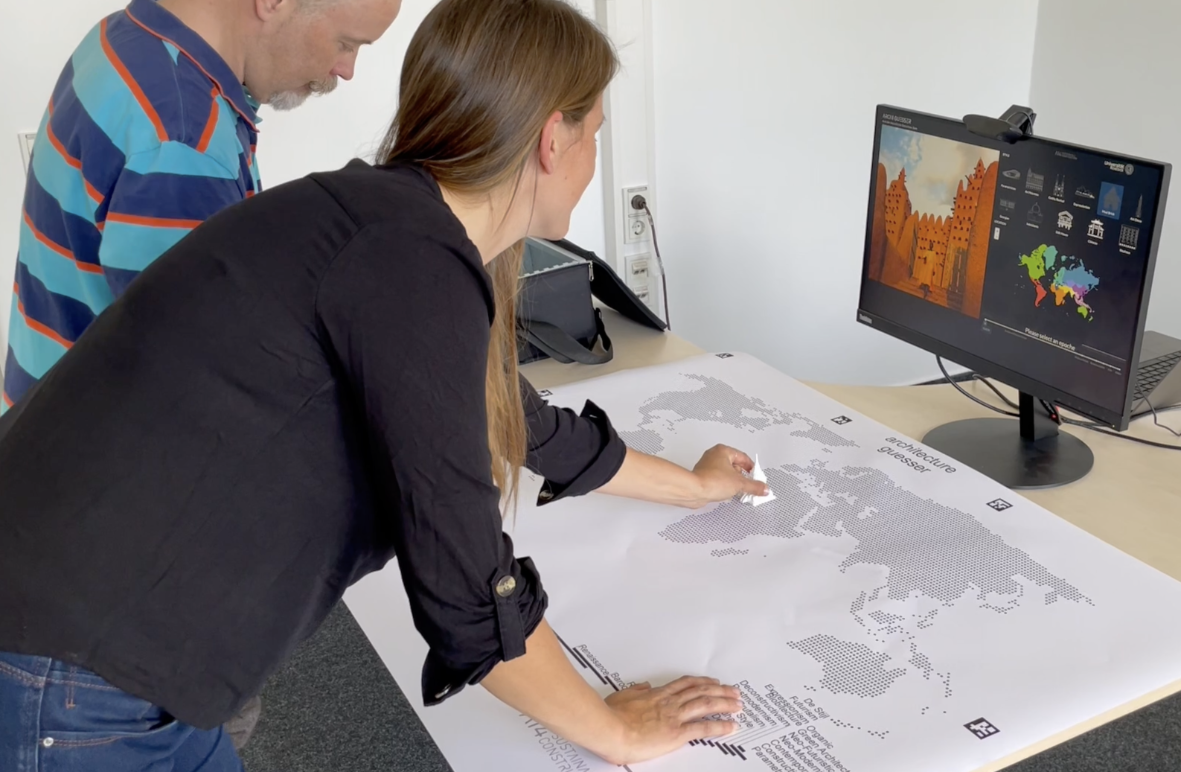}}
\subcaptionbox{Architecture \label{fig:architecture}}{\includegraphics[width=0.7\textwidth]{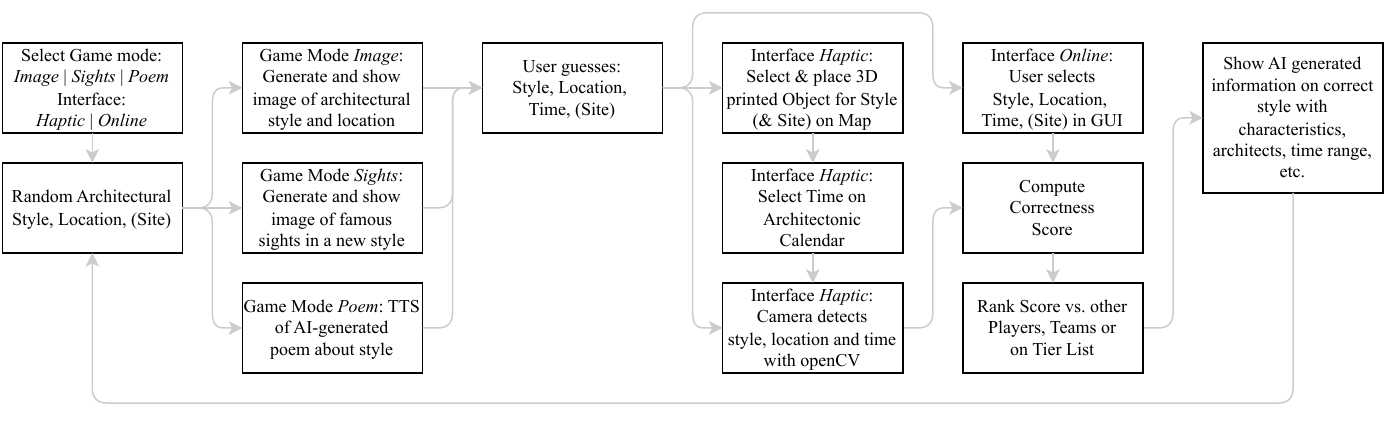}}
  \caption{ArchiGuesser Impressions}
  \label{fig:impressions}
\vspace{-1.5em}
\end{figure}

We designed the game to be multisensory to enhance memorization  \cite{quak2015multisensory, novak2021does} by generating visual and auditive content and providing a tactile interface in which the user gives his guess by positioning 3D-printed objects of the various architectural styles onto a map (see Fig.~\ref{fig:gameplay}). These 3D objects have been designed to embody distinctive style features giving the user visual and tactile hints for identification. The user is awarded points based on the proximity of his guess to the correct values and is ranked on a leaderboard. This makes the game enjoyable individually or in a group.

\section{The Generative AI pipeline}
We developed the game with the goal to create a multisensory experience that caters to different learning channels to teach architectural styles and also helps students understand the capabilities and workings of generative AI. Therefore, we decided that the content for the game should be generated primarily by AI including style descriptions, images, and poems. 

Figure~\ref{fig:architecture} shows the resulting \textit{ArchiGuesser} pipeline that is integrating various AI technologies. After selecting a game mode the game is presenting generated images or poems for a randomly selected style and optional landmark. The user's task is to identify the architectural style, geographical location, historical period, and landmark by positioning 3D-printed objects on a map. A physical slider allows users to specify the time period. We employ marker-based computer vision with OpenCV to identify the object types and positions. Subsequently, we compute a score to evaluate the user's accuracy in comparison to the ground truth. We present this score along with information about the correct style, providing users with additional information to learn to recognizing the styles.

In the game, we wanted to represent the richness of architectural history worldwide and avoid cultural bias. To achieve this, we utilized ChatGPT (GPT 4) to provide us with a list of the top 10 architectural styles for each socio-cultural region of the world, as outlined for example in \cite{regions2019} and \cite{zhai2010ancient}. This process was repeated 10 times. We selected the most consistent 30 responses to eliminate hallucination. Next, we tasked ChatGPT with providing concise JSON summaries encompassing style, time period, characteristics, and notable architects. With control questions we evaluate the quality of the response. Lastly, we meticulously curated the content to ensure accuracy, making only a few corrections as ChatGPT shows a solid knowledge of architectural history\cite{ploennigs24}.

We then generate images with Midjourney with two methods depending on the game mode. In the 'Image' mode, an image of a building in a specific architectural style is generated. We use templated prompts, such as "Building of <name> architectural style by <architect>," as well as prompts generated by ChatGPT that summarize the style using nouns and adjectives.
In the 'Sights' game mode, we have curated a selection of renowned local sites that exemplify distinct architectural styles. These images are uploaded to Midjourney for reinterpretation into a different style.

The poem game mode offers an alternative non-visual experience in which an architectural style is conveyed through recited poetry. We create these poems using templated prompts in ChatGPT that request a style-themed poem (describing opt.\ a landmark) without explicitly naming the style. To bring these poems to life, we convert them into speech using a custom-trained voice on elevenlabs.io.

\section{Conclusion}
Generative AIs offer us new and fun ways to reshape education and teach in engaging and creative ways. We present an example for a generative AI-driven multisensory educational game that teaches the diversity of architectural history, from the styles we see around us every day to unfamiliar styles from across the world.
With generative AI we can also break up these traditional cultural boundaries and remix them in new, fictional ways. We can show what Neuschwanstein might look like if built in the Chinese imperial era or what happens if Art Deco is fused with elements from African Mud Brick style.
This enables students to think about stylistic elements outside of their cultural bubbles. They can learn about the history of architectural styles, and at the same time gain new impulses for their own architectural designs. With generative AI all these novel possibilities can be made part of a simple and fun game like \textit{ArchiGuesser} (Demo video: \url{https://youtu.be/zuvKQeuoWuQ}).

\bibliographystyle{plain}

\end{document}